\documentclass[lettersize,journal]{IEEEtran}
\usepackage{amsmath,amsfonts}
\usepackage{algorithmic}
\usepackage{algorithm}
\usepackage{array}
\usepackage[caption=false,font=normalsize,labelfont=sf,textfont=sf]{subfig}
\usepackage{textcomp}
\usepackage{url}
\usepackage{verbatim}
\usepackage{graphicx}
\usepackage{cite}
\usepackage{indentfirst}
\usepackage{multirow}
\usepackage{hyperref}
\usepackage{xcolor}
\usepackage{adjustbox}
\usepackage{dblfloatfix}
\usepackage{fancyhdr} 

\usepackage{fancyhdr}

\begin{document}

\title{Unmasking Illusions: Understanding Human Perception of Audiovisual Deepfakes}

\author{Ammarah Hashmi, Sahibzada Adil Shahzad, Chia-Wen Lin,~\IEEEmembership{Fellow,~IEEE}, Yu Tsao,~\IEEEmembership{Senior Member,~IEEE}, and Hsin-Min Wang,~\IEEEmembership{Senior Member,~IEEE}
\thanks{Ammarah Hashmi is with the Social Networks and Human-Centered Computing Program, Taiwan International Graduate Program, Institute of Information Science, Academia Sinica, Taipei 11529, Taiwan, and also with the Institute of Information Systems and Applications, National Tsing Hua University, Hsinchu 300044, Taiwan. (e-mail: hashmiammarah0@gmail.com).

Sahibzada Adil Shahzad is with the Social Networks and Human-Centered Computing Program, Taiwan International Graduate Program, Institute of Information Science, Academia Sinica, and also with the Department of Computer Science, National Chengchi University, Taipei 11529, Taiwan. (e-mail: adilshah275@iis.sinica.edu.tw).

Chia-Wen Lin is with the Department of Electrical Engineering and the Institute of Communications Engineering, National Tsing Hua University, Hsinchu 300044, Taiwan. (e-mail: cwlin@ee.nthu.edu.tw).

Yu Tsao is with the Research Center for Information Technology Innovation, Academia Sinica, Taipei 11529, Taiwan. (e-mail:
yu.tsao@citi.sinica.edu.tw).

Hsin-Min Wang is with the Institute of Information Science, Academia Sinica, Taipei 11529, Taiwan. (e-mail: whm@iis.sinica.edu.tw).

}}



\maketitle
\thispagestyle{fancy} 

\begin{abstract}
The emergence of contemporary deepfakes has attracted significant attention in machine learning research, as artificial intelligence (AI) generated synthetic media increases the incidence of misinterpretation and is difficult to distinguish from genuine content. Currently, machine learning techniques have been extensively studied for automatically detecting deepfakes. However, human perception has been less explored. Malicious deepfakes could ultimately cause public and social problems. Can we humans correctly perceive the authenticity of the content of the videos we watch? The answer is obviously uncertain; therefore, this paper aims to evaluate the human ability to discern deepfake videos through a subjective study. We present our findings by comparing human observers to five state-of-the-art deepfake detection models. To this end, we used gamification concepts to provide 110 participants (55 native English speakers and 55 non-native English speakers) with a web-based platform where they could access a series of 40 videos (20 real and 20 fake) to determine their authenticity. Each participant performed the experiment twice with the same 40 videos in different random orders. The videos are manually selected from the FakeAVCeleb dataset. We found that all AI models performed significantly better than humans when evaluated on the same 40 videos. The study also reveals that while deception is not impossible, humans tend to overestimate their detection capabilities. Our experimental results may help benchmark human versus machine performance, advance forensics analysis, and enable adaptive countermeasures. 
\end{abstract}


\section{Introduction}
\IEEEPARstart{D}{eepfakes} are computer-generated or manipulated multimedia content that convincingly simulates or conceals reality. The emergence of Artificial Intelligence (AI) and Deep Learning (DL) technologies has advanced various fields \cite{R71, R72, R69, R70, R73} also manipulation tools that are capable of altering a person's identity by swapping voice or facial features. Such fabricated multimedia content, generated or manipulated through cutting-edge AI techniques, can depict a person in situations they've never been in. Deepfake generation technology is a gold mine with wide applications \cite{R62}; however, its dark side poses a significant threat to society and can have shocking consequences for unsuspecting citizens. For instance, it can cause chaos when highly realistic videos of public figures or politicians are generated to spread misinformation \cite{R5,R61 ,R63}. Fake videos can damage someone's reputation and emotional well-being by featuring them in explicit content without their consent \cite{R64}. The widespread malicious and unethical use of synthetic fake content on social media platforms has caused public unrest and raised serious trust issues in the media \cite{R5}\cite{R63}.

\begin{figure}[!t]
\centering
\includegraphics[width=3.3in]{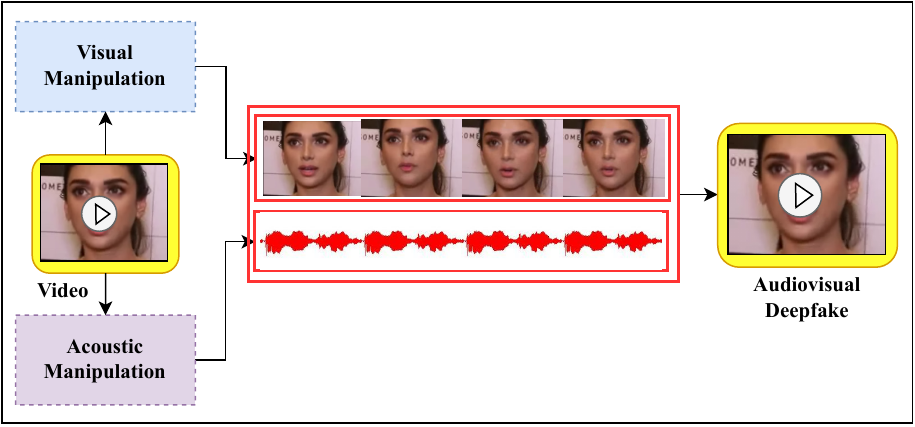}
\caption{An audiovisual deepfake example showcasing the simultaneous manipulation of audio and visual tracks to produce a highly convincing and deceptive deepfake video.}
\label{Fig1}
\vspace{-6mm}    
\end{figure}

Exponential advances in hyper-realistic visual effects and the manipulation across various modalities, including audio, image, and video, have led to the creation of increasingly plausible content, thereby undermining the trust in the proverb ``Seeing is believing'' in digital media. This raises concerns that audiovisually manipulated videos become visually and aurally indistinguishable. Audiovisual deepfakes are a form of synthetic media generated via sophisticated deep learning techniques that manipulate the acoustic and visual tracks of a video, often created with the intention of deceiving people. Fig. \ref{Fig1} provides an illustrative example of audiovisual deepfake creation. The development of neural generative models, such as Generative Adversarial Network (GAN) \cite{R6}, Autoencoder (AE), Variational Autoencoder (VAE) \cite{R7}, and diffusion model \cite{R67, R66, R68}, as well as multimedia content generation techniques based on them~\cite{R68, R51,R52,R53,R54,R55}, and the growing public accessibility of these models and techniques have had a strong impact on visual and auditory content. This highlights the importance of caution within the research community as it can be difficult to distinguish between truth and fabrication, creating new social challenges. In this paper, we study human perception of audiovisual deepfakes. Our contribution provides a deeper understanding of human performance in detecting audiovisual deepfakes, and our analysis may be valuable for enhancing cybersecurity measures, advancing forensic analysis, and developing countermeasures.   

Existing deepfake detection methods fall into two primary categories, unimodal \cite{R10, R30, R74} and multimodal \cite{R11, R13, R14, R15} deepfake detectors, each with specific strengths and weaknesses. Unimodal detectors, which exclusively focus on analyzing a single modality such as visual or audio, have made a substantial impact in establishing the basics of deepfake detection. Their strengths include simplicity and efficiency, focused analysis and low data requirements compared to multimodal approaches. Despite their better generalization ability, these methods are ineffective in the detection of advanced multimodal forgeries, audio deepfake detection models often miss visual manipulation, and visual deepfake detection models fail to detect audio manipulation. Multimodal detectors analyze multiple modalities such as audio and visual data streams to enhance the detection accuracy and robustness compared to unimodal approaches. However, multimodal deepfake detectors are resource-intensive or computationally expensive as they process and analyze both visual and audio tracks concurrently. Multimodal detectors can effectively detect known types of deepfakes, yet they suffer from limited generalization and may face challenges with novel or previously unseen manipulation techniques.

The generation \cite{R6, R16,R26,R10} and detection \cite{R10,R11,R13,R15,R14, R56, R57} of deepfakes have been extensively explored by the research community. While human perception of deepfake speech, images, or videos respectively has been a major focus of researchers, to date, there has been less research in the field of human perception of audiovisual deepfakes. This study is an empirical investigation designed to evaluate the ability of human participants to detect audiovisual deepfake videos. The objective of the experiment is to analyze the perceptual boundaries of audiovisual deepfakes and explore how the average human observers perceive and interpret fabricated content. Therefore, we conducted a comprehensive crowdsourced subjective test via a web-based platform, allowing human participants to watch a series of videos and identify authenticity. Evaluating human's ability to detect deepfakes is crucial, but it is also important to understand how detection algorithms identify them. Therefore, we also tested five state-of-the-art (SOTA) AI deepfake detectors on the same set of videos. Finally, we compared the subjective test results with the test results of the AI detection models. In particular, we explored the following questions:

\begin{itemize}
    \item Are participants able to differentiate between real and audiovisual deepfake videos above chance levels?
    \item Are participants better than SOTA AI algorithms at identifying audiovisual deepfake videos?
    \item After perceiving a video as a deepfake, are participants able to correctly identify specific forms of manipulation (i.e., audio and video)?
    \item Does familiarity with and forewarning of deepfakes improve participants’ detection accuracy?
    \item Does participants' self-reported level of confidence in their answers align with their accuracy in detecting audiovisual deepfake videos?
    \item Do participants' age, native language, and affinity for technology influence the perception of deepfake videos?
    \item Do participants who undergo several rounds of training show a better ability to identify deepfakes? 
\end{itemize}

The remainder of the paper is organized as follows. Section II briefly reviews related work. Section III outlines the details of the experiment, including the setup, design, dataset, and AI models. Section IV presents the human test results and compares them with the results of the AI models. Section V discusses the limitations of this subjective study. Finally, Section VI concludes the paper and proposes future work.

\section{Related Work}
Although many models have been proposed for detecting audio, image, video, and audiovisual deepfakes, in this section we only review related research on human perception of deepfakes.

In recent years, speech synthesis technology has made significant progress in producing speech that closely mimics human speech patterns and intonation~\cite{R16, R26}. Human perception studies conducted from different perspectives have shown that listeners often have difficulty distinguishing synthesized speech from natural speech, highlighting the effectiveness of this technology in creating highly realistic speech. In~\cite{R17}, experimental results in English and Mandarin highlight the difficulty for humans to identify audio deepfakes, with only 73\% of audio deepfakes correctly identified regardless of language. As speech synthesis algorithms continue to improve, detecting audio deepfakes may become more challenging. The study in~\cite{R18} compared human performance with AI models, showing that humans and deepfake detection algorithms exhibit the same strengths and weaknesses, but both have difficulty identifying certain types of attacks. Recent work in~\cite{R29} assessed human perception abilities by engaging participants in scenarios using synthetic voice messages under the guise of testing the user-friendliness of voice messages. This study showed that people have difficulty perceiving deception and that no one reacts to deceptive deepfake messages. \cite{R37} presented a comparative analysis between machine and human perception of adversarial speech, specifically synthetic audio commands that are recognized as valid messages. Meanwhile, the study in~\cite{R3} conducted an in-depth assessment of college students' perception of audio deepfakes, including how their academic majors and backgrounds influence their perceptions. 

For images, research on deepfake technology has shown that subtle changes in faces can generate compelling images that are difficult to detect by human observers \cite{R65}. Therefore, numerous studies have attempted to examine human perception capabilities to identify fabricated images, and have shown that human performance is not significantly better than the random chance level. For example, the subjective study in \cite{R32} analyzed the human ability to detect forgery in digital images, and found that humans can be easily fooled and that participants not only performed poorly at identifying image forgeries but also often doubted the authenticity of real images. The studies in \cite{R20, R43, R50} evaluated the human ability to identify image deepfakes of human faces, and found that people are not naturally good at detecting deepfakes and that the simple interventions tested do not help. \cite{R21} focused on a crowdsourcing-based approach to evaluate synthetic images specifically generated with GANs. The study in \cite{R27} shows that synthetically generated faces are not just highly photorealistic, they are nearly indistinguishable from real faces and are judged more trustworthy. The study in \cite{R38} explored the impact of synthetic image deepfakes on college students' perception and understanding of the technology. The study in \cite{R48} compared the ability of humans and machine learning models to detect synthetic images, with misclassification rates of 38.7\% and 13\%, respectively.

There are also subjective studies analyzing human ability to distinguish between authentic and synthetic videos, and comparing human performance with AI models. The study in~\cite{R23} shows that both human and AI models exhibit comparable performance, but with different errors, and that subjects with access to model predictions perform better than subjects without access to model predictions. The study in~\cite{R24} shows that while human perception is very different from machine perception, both are successfully fooled by deepfakes but in different ways. Specifically, algorithms have difficulty detecting deepfake videos that are easily spotted by humans. According to~\cite{R25}, people cannot reliably detect deepfakes. Raising awareness and financial incentives do not improve people's detection accuracy. People tend to mistake deepfakes as authentic videos (rather than vice versa) and overestimate their own deepfake detection abilities. Therefore, several studies have attempted to analyze the factors that influence viewers’ ability to perceive deepfake videos. The authors of~\cite{R1} conducted an investigative user study and analyzed existing AI detection algorithms to uncover the unknown factors behind the detection of deepfakes. A similar study in~\cite{R2} shows that two-thirds of participants were unable to accurately detect a sequence of four videos as either genuine or deepfake, and that familiarity with the subjects in the videos had a statistically significant impact on the individuals' perception ability. In addition, the studies in \cite{R44, R46} investigated psychological and social factors that influence people's ability to detect fake videos. \cite{R47} and \cite{R45} also contain interesting subjective studies, with the former investigating the role of system-generated cues, such as video quality and technical imperfections, and the latter focusing on developing and testing strategies to enhance human's ability to identify deepfakes. Another study in~\cite{R40} delved into how humans and machines perceive deepfake videos, suggesting that incorporating emotional factors and leveraging specialized visual processing may be promising strategies to enhance deepfake detection. The study in \cite{R41} explored the difference in detection performance between human observers and automated systems by comparing their abilities to identify deepfake videos in the presence of noisy channels.  The interesting work in~\cite{R4} examined tourists' visit intention by watching deepfake destination videos, and found that the factors that affect the tourists' visit intention after watching deepfake videos include information manipulation tactics, trust and media richness, while perceived deception does not influence tourists' visit intention. The study in~\cite{R5} shows that people are more likely to feel uncertain than to be misled by political deepfakes, but this resulting uncertainty, in turn, reduces trust in news on social media. The authors of~\cite{R42} explored the impact of deepfake videos' informative cues on individuals' perceived accuracy of claims and their intentions to share non-political deepfakes, and found that high cognitive individuals were less likely to trust news and share content, and that people were more likely to believe that deepfake claims were true when informative cues were absent. The results in~\cite{R9} indicate various challenges in deepfake user perception that technology developers need to address before the potential of deepfake applications can be realized for human-computer interaction.

In addition to the above-mentioned human perception and cognition of single-modal synthetic (or manipulated) speech, images, and videos, there have also been a few studies exploring cross-modal human perception. For example, the authors of~\cite{R39} conducted a comprehensive survey of 3,002 participants across three countries to understand humans' ability to detect synthetic media across audio, image, and text. The results show that SOTA forgeries are nearly indistinguishable from ``authentic'' media, highlighting the difficulty of identifying deepfakes. Similarly, the authors of \cite{R34, R49} studied the effectiveness of human discernment in identifying AI-generated content across various media formats, including image, video, audio, and audiovisual stimuli. The authors of \cite{R49} studied the psychological and perceptual effects on people after being deceived by deepfakes. The study in~\cite{R22} shows that individuals are more vulnerable to, and more likely to share, video deepfakes than other forms of deepfakes (e.g., audio deepfakes and cheapfakes\footnote{Cheapfakes, also known as shallowfakes, are media content manipulated using non-AI methods, such as basic video/audio editing tools.}), and that individuals with high cognitive ability are less likely to perceive deepfakes as accurate or share them. Another experimental study in~\cite{R19} compared the performance of human perception in detecting political deepfake speeches across text, audio, and video, finding that audio and visual information enables more accurate discernment than text alone.

Research on human perception of audiovisual deepfakes is emerging but remains relatively underexplored. For example, the study in~\cite{R23} shows that when synthesized speech is subtle and well synchronized with corresponding lip movements, audiovisual deepfakes can achieve a high level of realism that challenges human perception. The authors of \cite{R8} conducted a pilot study on the impact of physical attractiveness on students' evaluations of teachers and found that students' ratings of the quality of teachers' deepfake videos and real videos were comparable, and that students did not detect deepfake. The authors of \cite{R28} conducted a subjective evaluation of media consumers' vulnerability to fake audiovisual content, and found that their detection performance improved when participants knew the displayed individual or when a biometric reference video (introducing the individual and its behavior) was available to them during testing.
As technology advances and manipulated content becomes more common, so does the potential for audiovisual deepfakes to deceive individuals and influence public opinion. To understand how humans perceive and interpret incredibly realistic audiovisual deceptive content, our study adopts a subjective approach that allows for a nuanced understanding of how individuals perceive and evaluate audiovisual deepfakes, shedding light on the cognitive processes involved.

\begin{figure}[!t]
\centering
\includegraphics[width=3.0in]{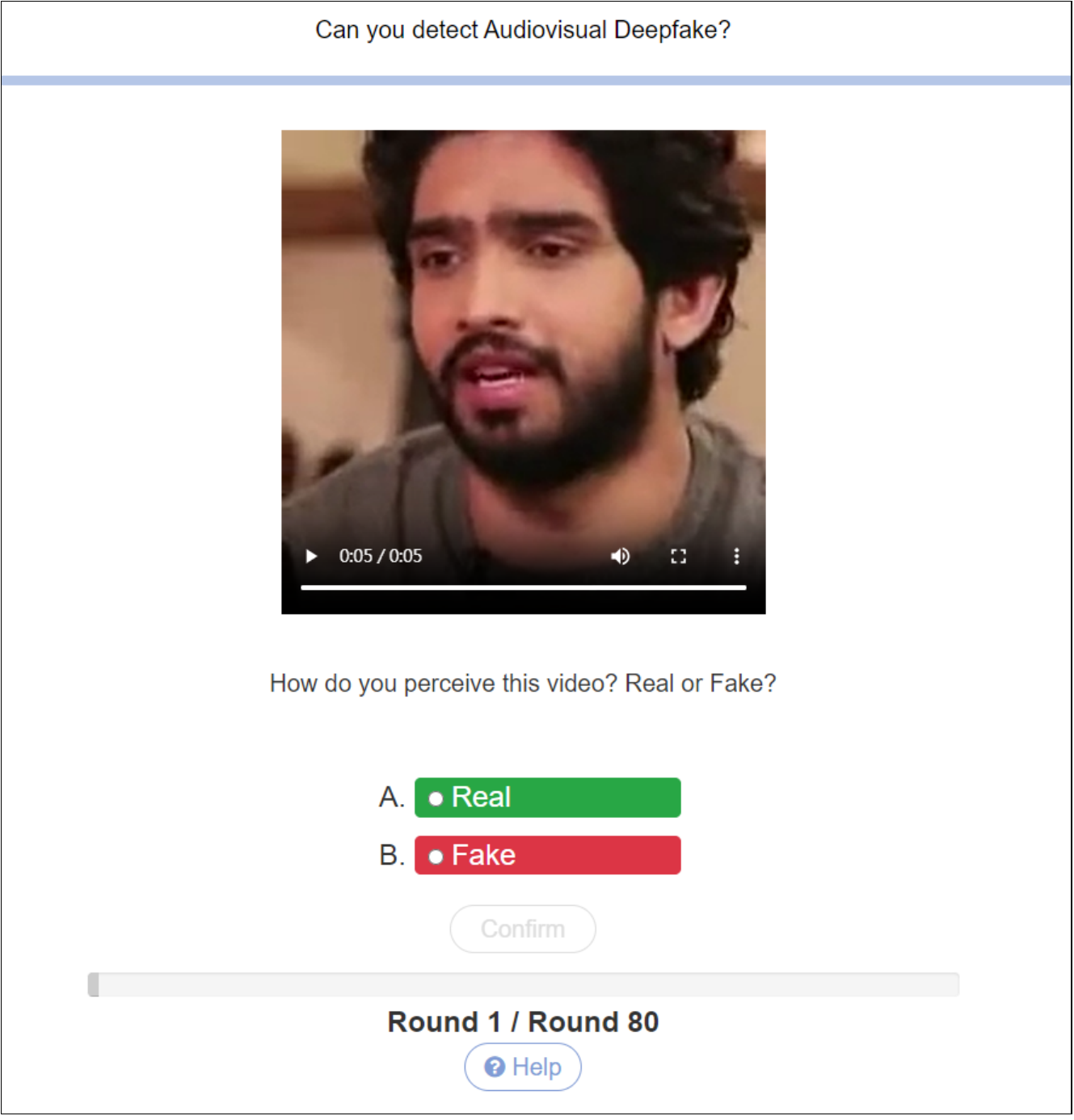}
\caption{The web interface where participants watch videos and identify whether the videos are 'Real' or 'Fake'. After classification, participants are shown the labels of the videos. The web interface can be accessed via \url{https://jack24.mooo.com/web/experiment/frontend/web/}
}
\label{Fig2}
\vspace{-6mm}
\end{figure}

\section{Experiment}
We conducted an online experiment among English-speaking participants with the primary aim of determining the extent to which synthetically generated audiovisual content misleads viewers or creates uncertainty about the authenticity of the video. In other words, we developed an online platform to analyze human performance and perception of fake videos. Our study delves into human behavior, exploring the psychological and perceptual factors that lead people to be misled or experience uncertainty when exposed to content that include acoustic and visual manipulations. This section outlines the structure of the experiment, the dataset we used, the sample selection process, and how we curated the samples presented to participants. Additionally, we discuss five AI detection models for comparison with human performance. The focus of our experiment is to evaluate the potential of humans in detecting audiovisual deepfakes, not to evaluate AI detectors.

\subsection{Experiment Description}
We organized an online experiment and recruited subjects through the crowdsourcing platform Amazon Mechanical Turk. Each participant was paid \$10 to complete the study, which typically lasted an average of 30 minutes. We provided the participants with the URL to conduct the experiment and informed them of the eligibility criteria for participating in the study. Each participant was assigned a unique ID. Participants must retain this unique ID for use in the payment process. When submitting their task on Amazon Mechanical Turk, participants were required to enter this unique user ID as proof of completion. Failure to provide a valid user ID will result in ineligibility for payment processing. This protocol ensures the integrity of the payment process and helps identify and exclude incomplete or fraudulent submissions.

\subsubsection{Participant Selection}
The participants for the study were recruited through both convenience sampling and targeted recruitment, ensuring diverse representation based on age, gender, and background. There were 110 participants in the study, 
and participation criteria include:

\begin{itemize}
    \item Aged between 20 and 50 years old.
    \item No history of visual or hearing impairment that might affect the ability to evaluate audiovisual content.
    \item Beginner, intermediate, good, or expert level  understanding of digital media or synthetic media technology.
\end{itemize}
Before participating in the experiment, participants were informed of the purpose of the study and asked to provide informed consent.

\subsubsection{Study Procedure}

Our experiment presented a binary classification challenge, in which participants were shown a series of video clips, one at a time, and had to discern whether the video was real or manipulated, as shown in Fig. \ref{Fig2}. Participants were informed that there was a 50\% chance that each video was fake and encouraged to rely strictly on their own instincts and observations.

To assess their ability to identify the source of manipulation in situations where participants thought the video was fake, they were prompted to carefully specify which modality of the video was altered -- video, audio, or both. In the case of video manipulation, participants were further asked to indicate which aspects of the visual manipulation were noticed, such as whether the lips, face, or eyes were manipulated or the audio and video were out of sync. Likewise, in the case of audio manipulation, participants were asked to choose between the audio being manipulated or being out of sync with the video. Once participants classified a video, they were provided with the correct label; in this way, participants could track their performance. It is worth noting that the correct label is either ``real'' or ``fake'', excluding specific manipulation types in each modality. Each participant had to complete 80 rounds, i.e., judging 80 videos. Our experiment consists of two phases. Participants judged 40 videos presented in a random order in the first phase, and then again judged the same 40 videos presented in another random order in the second phase. This was to see if individuals performed consistently the second time they judged the videos.
Upon completion of the 80-rounds experiment, participants were asked to indicate their level of confidence. Each participant was asked to indicate how confident they were in the accuracy of their answers.
In the end, participants received feedback on the number of correctly classified samples out of the total number. 

\begin{figure*}[!t]
\centering
\includegraphics[width=7.0in]{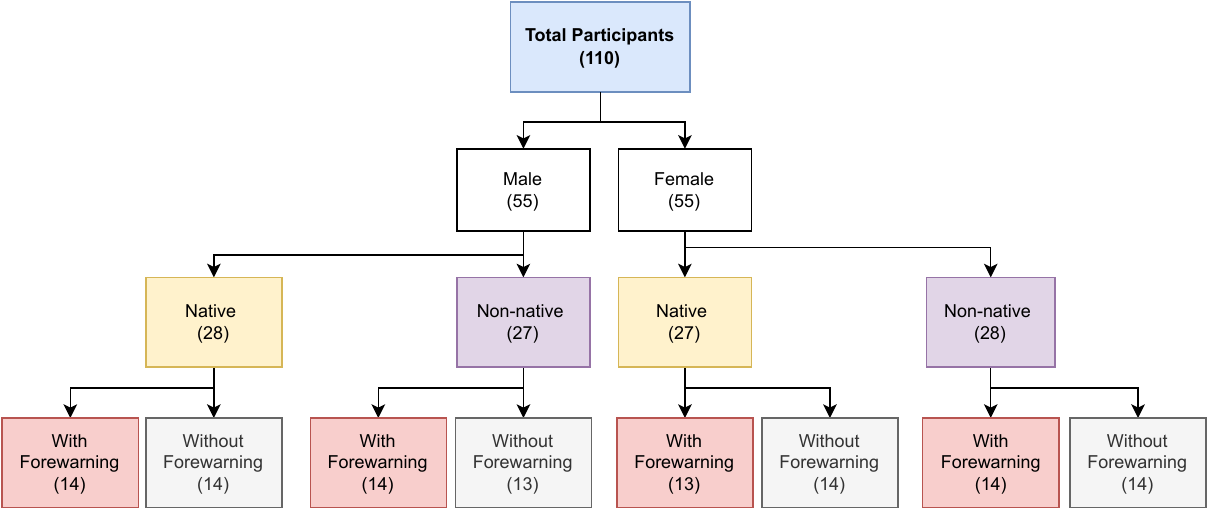}
\caption{Distribution of participants in the experiment.} 
\label{Fig3}
\vspace{-4mm}
\end{figure*}

\subsection{Participants}
\begin{figure}[!b]
\vspace{-8mm}
\centering
\includegraphics[width=3.0in]{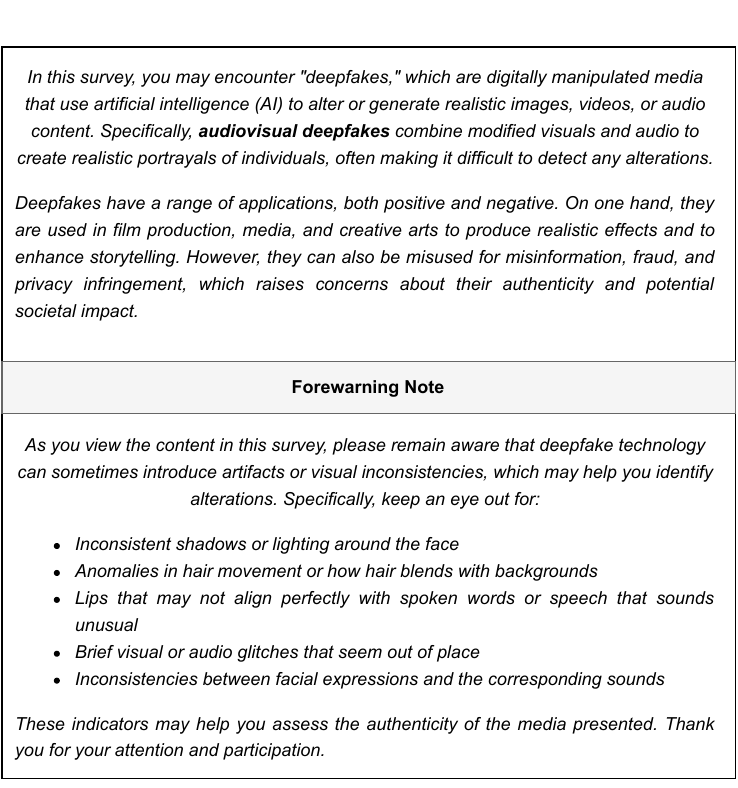}
\caption{Forewarning note presented to half of the participants at the beginning of the experiment.}
\label{Forewarning}
\end{figure}

We excluded participants who participated in less than 80 rounds. A total of 110 participants completed the experiment, evenly distributed by gender, with half being native English speakers and the other half being non-native English speakers. Half of the participants received a forewarning note, as shown in Fig. \ref{Forewarning}, about the threat of deepfakes, while the other half did not. This forewarning note informed participants about audiovisual manipulation, highlighted the threats posed by deepfakes, and further emphasized the need to be aware of potential visual irregularities such as inconsistencies between the speaker’s facial movements and the audio, unnatural lip movements or glitches in videos. This setting allowed us to explore whether prior awareness and understanding of the risks associated with deepfakes impacted participants' performance. 

\begin{figure}[!b]
\vspace{-6mm}
\centering
\includegraphics[width=3.0in]{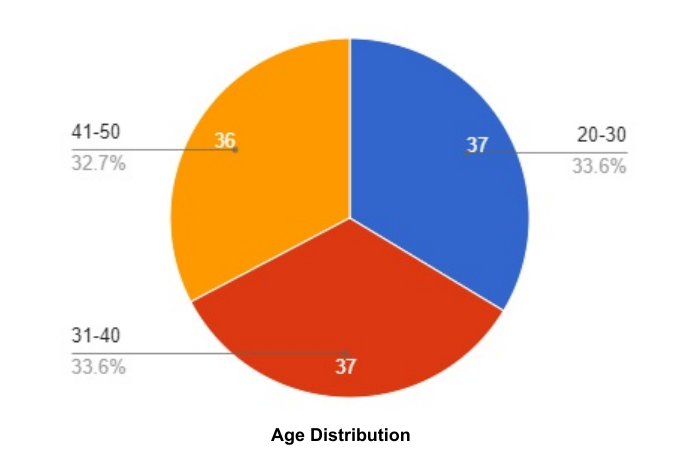}
\caption{Age distribution of participants in the experiment.}
\label{Fig4}
\end{figure}

The overall distribution of participants is shown in Fig. \ref{Fig3}. The age distribution of participants is shown in Fig. \ref{Fig4}. Additionally, we asked participants about their information technology (IT) skill level by providing four options: beginner, intermediate, good, or expert. This distribution is shown in Table \ref{table1}.

\begin{table}[!h]
\caption{Distribution of participants at different IT skill levels.}
    \centering
    \scalebox{1.1}{
    \begin{tabular}{|c|c|}
         \hline
         \textbf{IT Skill Level} & \textbf{Number of Participants} \\
         \hline \hline
         Beginner & 28 \\
         \hline
         Intermediate & 28 \\
         \hline
         Good & 27 \\
         \hline
         Expert & 27 \\
        \hline
    \end{tabular}}
    \label{table1}
\end{table}

\begin{figure*}[!t]
\centering
\includegraphics[width=7.0in]{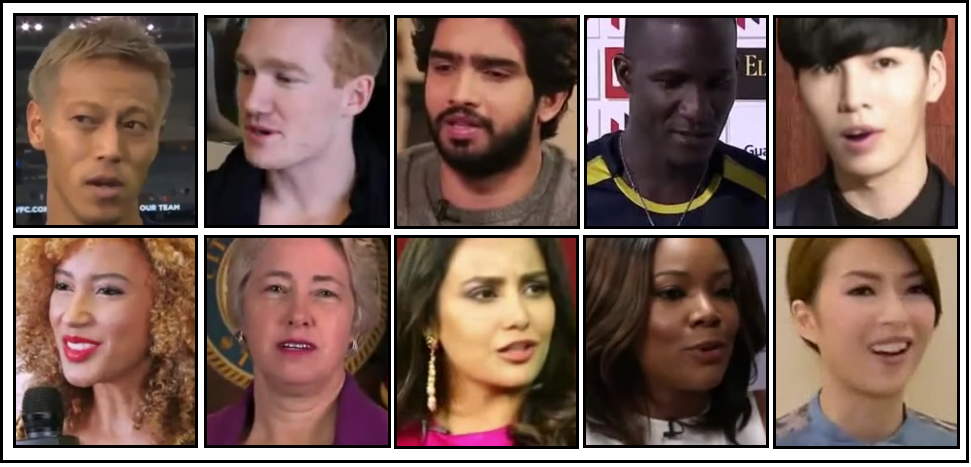}
\caption{Sample videos from the FakeAVCeleb dataset, including individuals of varying ages, ethnic backgrounds, and genders (faces cropped from the videos).}
\label{Fig5}
\vspace{-4mm}
\end{figure*}

\subsection{Dataset}
We conducted the study using the FakeAVCeleb dataset \cite{R12}, a dataset specifically designed for the task of audiovisual forgery detection. The dataset contains 500 videos featuring ethnically diverse celebrities from industries including music, sports, politics, and film, prioritizing diversity in ethnicity, age, geography, race, and gender. These 500 original videos were used to generate 19,500 fake video samples. The videos of 430 celebrities were used for training, and the videos of 70 celebrities were used for testing. The fake videos cover audio manipulation, video manipulation, or both, using advanced tools including SV2TTS \cite{R33} (a transfer learning-based real-time voice cloning tool) for synthetic clone voice generation and Wav2Lip \cite{R16}, Faceswap \cite{R35}, and Fsgan \cite{R36} for visual manipulation. The videos in the dataset are divided into 4 categories depending on whether they were acoustically or visually manipulated, namely Fake-Video-Fake-Audio (FVFA), Fake-Video-Real-Audio (FVRA), Real-Video-Fake-Audio (RVFA), and Real-Video-Real-Audio (RVRA). These detailed labels for audio and video streams facilitate training and performance evaluation of the forgery detection system. Fig. \ref{Fig5} shows some video examples.

We chose to use the FakeAVCeleb dataset in our study because of the following important factors: 1) video diversity, such as four ethnic backgrounds, various geographic locations, and both genders, 2) use of the latest manipulation techniques, 3) comprehensive labeling, and 4) multimodal manipulation, involving both audio and video manipulations, in line with our research focus. By leveraging this dataset, we aim to delve into human behavioral patterns in identifying audiovisual deepfake videos. Incorporating audio and visual manipulations into the dataset allows us to gain valuable insights into how individuals interpret and identify fakes across modalities.

\subsubsection{Data Selection}
In our experiment, we manually selected 40 videos from the test set of the FakeAVCeleb dataset, 20 videos from the real class, and 20 videos from the fake class. In each class, 10 videos feature males, and 10 videos feature females. No preprocessing steps were performed on the videos before they were used in the study.

For the fake class, we selected 4 videos for each gender from the FVFA and FVRA categories, respectively, and 2 videos for each gender from the RVFA category. Since the FakeAVCeleb dataset contains various manipulation types across acoustic and visual modalities, to keep the test set balanced, we selected samples from each manipulation type.

\subsection{Web Platform}
The experiment was conducted using our gamification-based web platform. We leveraged the popular open-source web server nginx to engage participants in the task of identifying the authenticity of videos they watched. For backend development, we used the high-performance framework PHP Yii2 and used PHP 7.4 as the developing language. We used the PHP $shuffle()$ function based on the Mersenne Twister algorithm to ensure repeatability and randomize the videos. MongoDB, known for its flexibility and performance, was used as a NoSQL database, with MongoDB Compass providing an intuitive graphical user interface (GUI). For efficient code editing, we used Sublime Text. 

\begin{figure*}[!b]
\vspace{-4mm}
\centering
\includegraphics[width=7.0in]{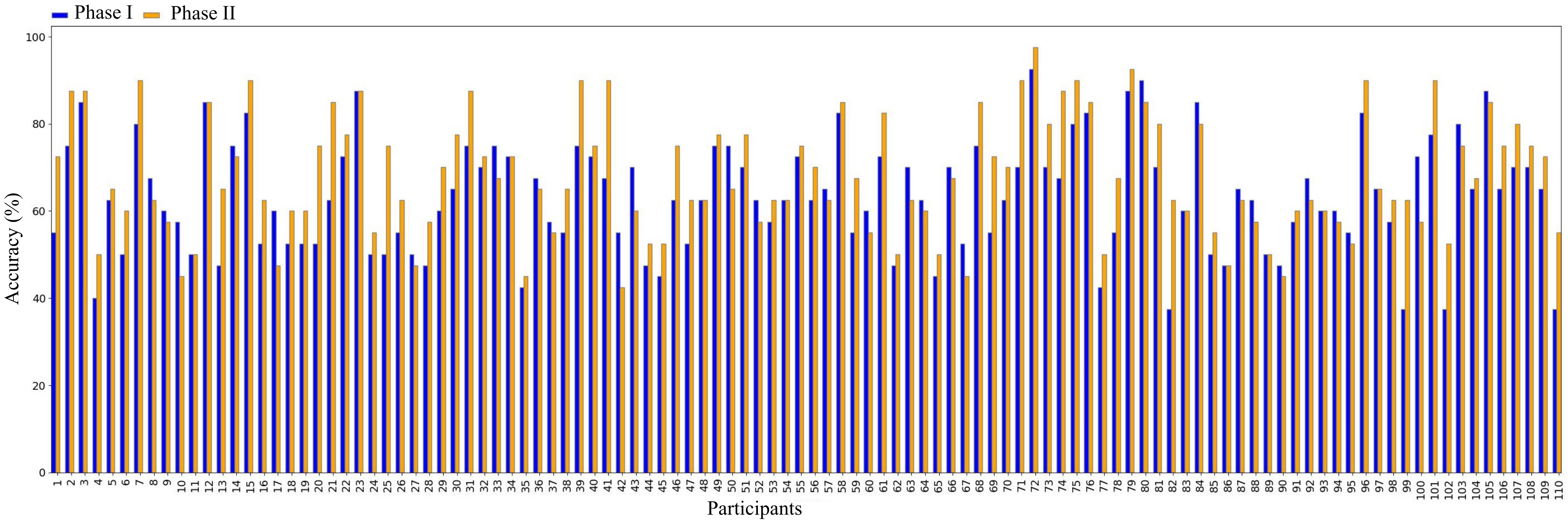}
\caption{The average detection accuracy of individual participants across all videos in Phases I and II.}
\label{Fig6}
\vspace{-6mm}
\end{figure*}

\subsection{Metrics}
We used accuracy as a metric to evaluate human perception of audiovisual deepfakes, which measures the overall correctness of participants' judgments in distinguishing real videos from deepfake videos. Accuracy refers to the proportion of correctly identified videos to the total number of videos presented to participants, and is calculated as follows:

\begin{equation}
\text{Accuracy} = \frac{\text{Number of Correctly Identified Videos}}{\text{Total Number of Videos}} \times 100\%.
\end{equation}

In conjunction with accuracy, we evaluated human perception of audiovisual deepfakes using a combination of qualitative analysis and contextual factors including gender, age, native language, and IT skill level. Qualitative analysis refers to the examination and interpretation of non-quantitative data, focusing on the qualitative aspects of participants' responses, behaviors, and experiences. 

\subsection{Gamification Concepts}

Participant motivation plays a crucial role in maintaining engagement and completion rates in subjective studies, especially when the tasks are extensive or repetitive. The risk of the task becoming boring after a few rounds poses a significant challenge that can strongly influence the design process of the study. We used several gamification concepts to keep participants motivated and engaged. Specifically, four approaches were employed. 

\begin{itemize}
\item Incentive (Monetary Reward): We offered a monetary reward or incentive of \$10 upon completion of the study. This approach not only serves as a powerful incentive for participants to participate but also contributes to a more effective and rewarding research experience overall. 

\item Points and Scoring: Participants received visual feedback showing the number of rounds completed and points earned through the progress bar and correct identification, thus encouraging continued engagement and motivation. In addition, we divided the task into two phases. This division allows participants to experience a sense of accomplishment after completing the first phase. By providing this tangible sense of accomplishment at the midpoint, participants are empowered to complete the second phase with renewed energy and focus. 

\item Progress Tracking: Participants could track their progress throughout the study and maintain their attention and engagement through visual feedback on task completion.  

\item Feedback Mechanism: We exposed participants to the correct label after each round. This immediate feedback keeps their interest and helps them track their performance. These strategies maintained engagement and ultimately increased their motivation to complete the study.
\end{itemize}

\subsection{AI Models}
We evaluated five SOTA deepfake detection models. The selection of these five AI deepfake detection methods for this study was based on their status as recent state-of-the-art models and the availability of their test results on the dataset we utilized. These models were trained on the training set of the FakeAVCeleb dataset and tested on the same test set used in the subjective evaluation. This comparison provides scientific insights into human and machine behavior when detecting audiovisual deepfakes.

\begin{itemize}
\item LipForensics \cite{R30} focuses on identifying significant semantic inconsistencies in mouth movements that are often present in AI-generated video content. Note that it is a visual-only model and cannot detect fake videos with only audio manipulation. 

\item AV-Lip-Sync~\cite{R15} utilizes speech-lip synchronization to detect forgery in videos. Specifically, it identifies inconsistencies between the lip movements observed in the video and the synthetic lip movements generated from the audio using the Wav2lip model. 

\item AV-Lip-Sync+~\cite{R11} is also based on speech-lip synchronization. It extracts acoustic and visual features via a transformer-based self-supervised learning (SSL) pre-trained AVHuBERT \cite{R31} model. 

\item CNN-Ensemble is a CNN-based ensemble network \cite{R13}, consisting of an audio-only network, a video-only network, and an audiovisual network. The outputs of the three networks are fused by majority voting.  

\item AVTENet is a transformer-based ensemble multimodal deepfake detector~\cite{R14}. The model consists of an audio-only network, a video-only network, and an audiovisual network, all of which are based on pre-trained transformer-based models for feature extraction. It has various fusion strategies. In the experiment, we used feature fusion. 

\end{itemize}

\begin{figure*}[!t]
\centering
\includegraphics[width=7.0 in]{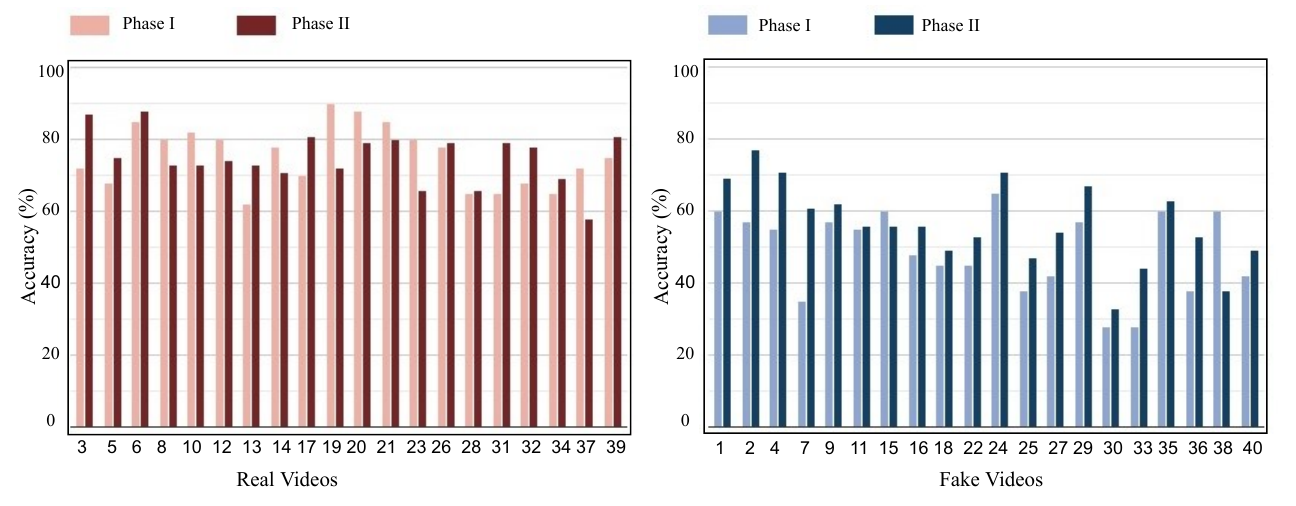}
\caption{The average detection accuracy for individual videos across all participants in Phases I and II.}
\label{Fig7}
\vspace{-4mm}
\end{figure*}

\section{Results}
In this section, we discuss the results of our subjective analysis, focusing on the performance comparison between human observers and AI algorithms based on various factors. We delve deeper into the efficacy of human perception in detecting audiovisual deepfakes to gain insights. The core findings of our work are summarized as follows:

\begin{itemize}
    \item Participants demonstrated an above-chance ability to differentiate between real videos and deepfake videos.
    \item Human observers performed worse than SOTA AI algorithms at identifying deepfake videos.
    \item Participants had difficulty identifying the specific form of manipulation after perceiving a video as fake, and tended to attribute it to visual manipulation even when acoustic manipulation was present.
    \item Familiarizing them with and being forewarned about deepfakes did not significantly improve participants' detection accuracy.
    \item Participant’s self-reported level of confidence did not align with their actual accuracy in detecting audiovisual deepfake videos.
    \item Participants' age, gender, and native language affected their perception of deepfake videos, but their affinity for technology did not.
    \item Participants showed an enhanced ability to discern deepfakes after several rounds of training.
\end{itemize}

\subsection{Overall Detection Accuracy across Participants}

Fig. \ref{Fig6} shows the Phase I (40 rounds) and Phase II (40 rounds) deepfake detection accuracy of individual participants. From Phase I to Phase II, the average accuracy across all participants increased from 63.30\% to 67.98\%, resulting in an overall accuracy of 65.64\%. Most participants performed better in Phase II than in Phase I.
Human performance at detecting audiovisual deepfakes is marginally better than random chance. This study demonstrates that deepfakes have the ability to deceive the majority of the public. This result is consistent with previous research \cite{R18, R20}, \cite{R48}, \cite{R24}.

\begin{figure}[!b]
\vspace{-8mm}
\centering
\includegraphics[width=3.5 in]{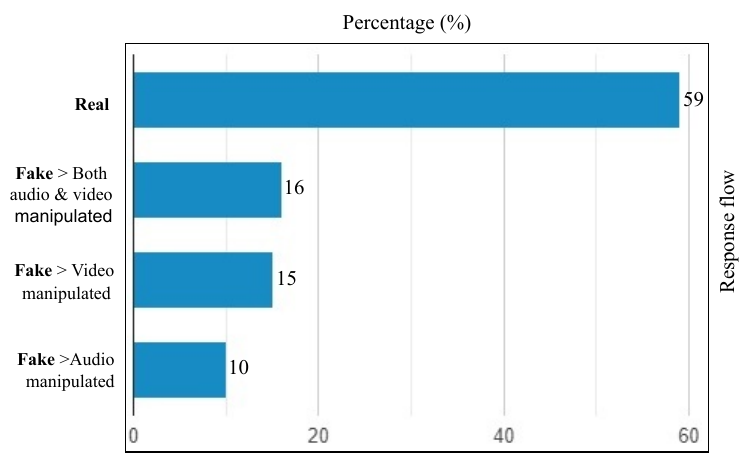}
\caption{Proportional distribution of participants' responses.}
\label{Fig8}
\end{figure}

\begin{figure*}[!b]
\centering
\includegraphics[width=7.0 in]{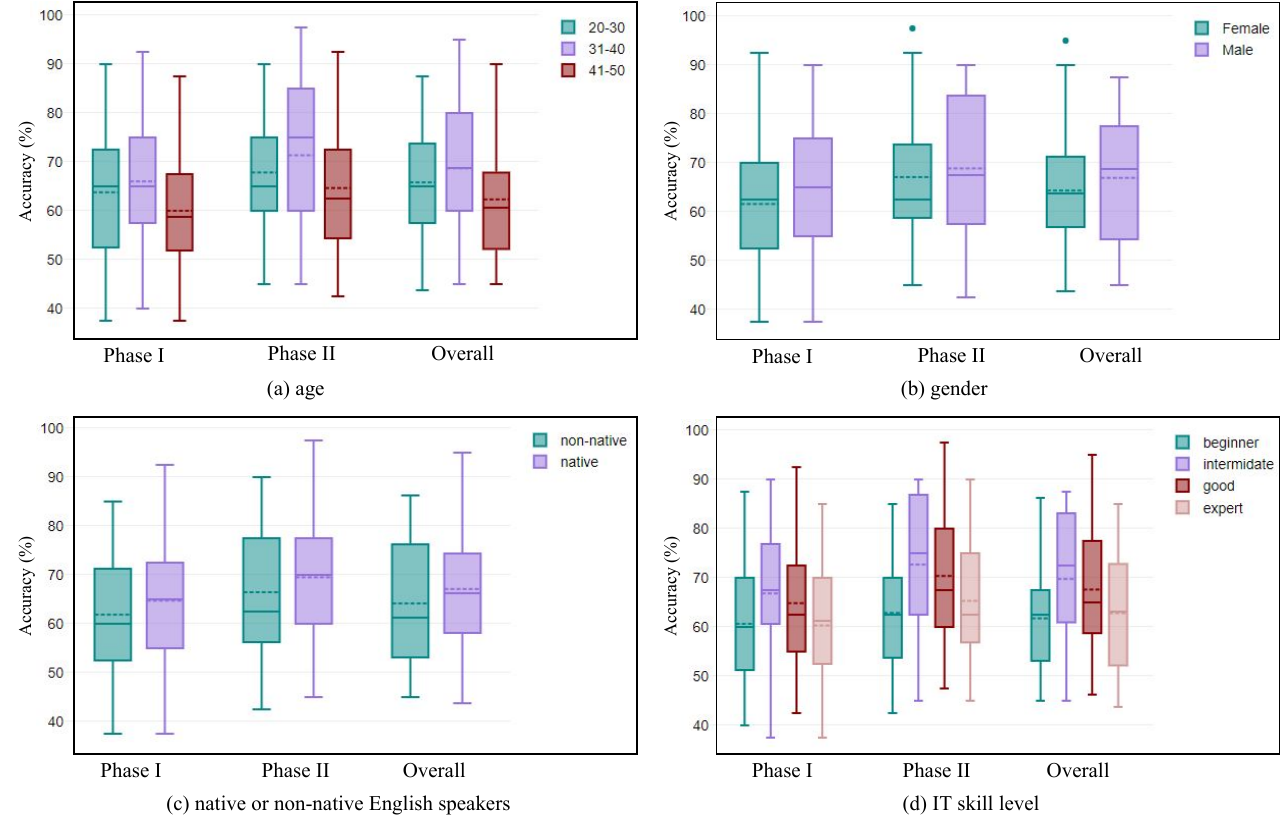}
\caption{Human detection accuracy grouped by (a) age, (b) gender, (c) native or non-native English speakers, and (d) IT skill level. The box in each box-and-whisker plot illustrates the interquartile  range (IQR), covering the middle 50\% of the data, while the upper and lower whiskers extending from the box show the range of the overall data. Data points outsides the whisker boundaries are outliers, representing unusually high or low values. The line within the box represents the median, and the dashed line within the box represents the mean of the data.}
\label{Fig9}
\vspace{-4mm}    
\end{figure*}

We performed a two-tailed t-test to see if the difference in average accuracy between Phase I and Phase II was significant. The significance level ($\alpha$) was set at $0.05$. The $t$-value is $-5.61$, and the corresponding $p$-value is $1.52 \times 10^{-7}$, which indicates a statistically significant difference in accuracy between the two phases.
\textcolor{red} {We hypothesize that people improve performance naturally through practice, familiarity, feedback, pattern exposure, or specific audiovisual cues. To prove this, we also performed a one-tailed t-test, as we hypothesized that the average accuracy of Phase II should improve compared to Phase I. The $t$-value remains $-5.61$, and the $p$-value is $7.62 \times 10^{-8}$. This extremely small p-value suggests support for the conclusion that the improvement from Phase I to Phase II is meaningful and statistically significant.} 

The results show that training is effective and that increased exposure and familiarity help participants learn and adapt to the stimuli. Over time, their anxiety may reduce and attention and concentration may increase. Furthermore, after each round, the participants were exposed to the correct video label. This immediate feedback on their performance reflected their correctness and errors, the former increasing their confidence in the detection, and the latter enhancing their ability to accurately distinguish real content from synthetic content. 

Fig. \ref{Fig7} shows the average detection accuracy for individual videos across all participants in Phases I and II. Most videos had better detection accuracy the second time they were shown to participants, especially fake videos. Notably, participants continued to correctly identify fabricated videos as deepfake; however, they might experience confusion and struggle to detect genuine videos as real. Nevertheless, despite the confusion, they still managed to perform better and show overall improvement in Phase II. This upward trend highlights the effectiveness of continuous learning to improve detection capabilities over time.

Interestingly, we also observed a consistent pattern of authenticity bias, with people being more biased towards detecting videos as real, even though they were well-informed that half of the videos were deepfake. This cognitive bias is evident in Fig.~\ref{Fig8}, where 59\% of videos were perceived as real and 41\% were identified as fake. Our findings suggest that people follow the principle of ``seeing is believing'' and tend to believe visual information, leading to interpreting a video as authentic even if it is not. This bias can have a significant impact on decision-making, as people may rely more on intuition than objective evidence, which can lead to
being deceived by fabricated content.

Furthermore, when participants were prompted to indicate the manipulated modality after identifying a video as fake, they tended to believe that both modalities (audio and video) were manipulated, or that there was only visual manipulation. This tendency may stem from visual primacy or cognitive load. Humans are highly visual creatures, and our brains prioritize visual stimuli over auditory signals. One potential factor may be sensory discrepancy; in some cases, deepfake videos exhibit more pronounced visual discrepancies than auditory ones. Furthermore, processing audio and visual information simultaneously may cause cognitive challenges; therefore, participants found it easier to focus on visual cues, encouraging them to prioritize visual information when detecting manipulations in videos. As shown in Fig. \ref{Fig8}, of the 41\% of ``fake'' responses, 16\% believed that there were two manipulation modes, 15\% believed that there was only visual manipulation, and only 10\% believed that there was only auditory manipulation. However, audio manipulation accounts for 10\% of all videos, and visual and audiovisual manipulation accounts for 20\%. The results show that humans tend to underestimate the actual number of fake samples in detection.


\subsection{Detection Accuracy under Different Participant Groupings}

In the following analytical study, we examined how the human ability to perceive audiovisual deepfakes correlates with four factors: age, gender, native or non-native English speakers, and IT skill level. For each factor, we aimed to determine important trends that might influence participants' effectiveness in identifying fakery in the video.

As shown in Fig. \ref{Fig9}(a), older participants (41-50 years old group) had lower accuracy compared to younger participants in the 20-30 years old and 31-40 years old groups. Differences in digital literacy, exposure to digital technology, and familiarity with digital media manipulation techniques may be the reason. Young people who have grown up in the digital age may be more familiar with media deception. As a result, they have a sharper eye for identifying deepfakes and experience in critically evaluating online content. In contrast, older participants may have difficulty identifying subtle signs of manipulation in deepfake videos due to their limited exposure to digital media manipulation. Another possible explanation for the lower detection accuracy in the older participants is that their cognitive abilities decline with age (e.g., cognitive abilities such as memory and attention may be affected), while younger individuals may have greater cognitive flexibility, which helps them to adapt to new information more quickly and distinguish truth from falsehood.

\begin{figure}[!b]
\vspace{-6mm}
\centering
\includegraphics[width=3.5in]{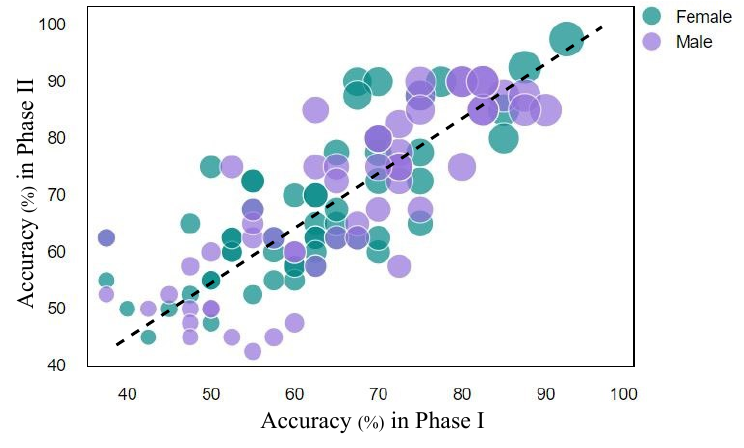}
\caption{Audiovisual deepfake detection accuracy of individual female and male participants in Phases I and II. Circle size indicates the overall accuracy of participants' judgments, with larger circles indicating higher accuracy.}
\label{Fig10}
\end{figure}

The analysis results in Fig. \ref{Fig9}(b) show that female participants are generally inferior to male participants in identifying the authenticity of videos. There may be several potential factors that contribute to this observed difference. For example, men, on average, are more interested in and exposed to technology and therefore more familiar with fabricated content. Stereotypes and social norms are also possible factors, with men more likely to engage in activities and interests involving technology and digital media manipulation. Cognitive differences between men and women may also affect their ability to detect deepfakes, such as differences in attention to detail or spatial reasoning skills. However, we found that female participants had better learning capabilities, with the average accuracy increasing from 61.59\% in Phase I to 67.09\% in Phase II, while male participants' performance improved less, from 65.00\% in Phase I to 68.86\% in Phase II. Fig. \ref{Fig10} reflect the results. Thirty-six of the 55 female participants performed better in Phase II than in Phase I, whereas only 33 of the 55 male participants performed better in Phase II than in Phase I. A higher proportion of female participants performed better in Phase II than in Phase I compared to male participants. Furthermore, Fig. \ref{Fig10} also shows that participants who were more accurate in Phase I also tended to be more accurate in Phase II, and vice versa.

\begin{figure*}[!t]
\centering
\includegraphics[width=7.0in]{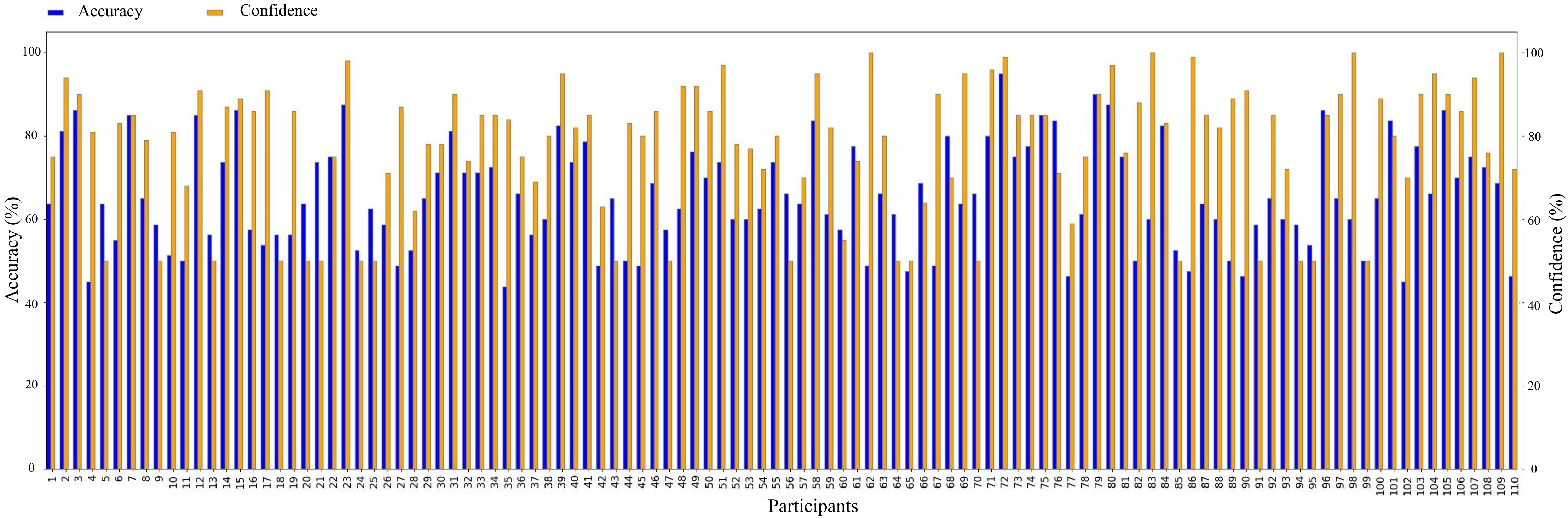}
\caption{Participants' actual detection accuracy and self-reported confidence in audiovisual deepfake detection. Participants' average self-reported confidence is 77.60\%, which is higher than the average accuracy of 65.64\%.}
\label{Fig11}
\vspace{-4mm}
\end{figure*}

From Fig. \ref{Fig9}(c), we can see that native English speaking participants performed better than non-native English speaking participants. This may be attributed to language proficiency, as our experiment was conducted on English-speaking videos. Native English-speaking participants tended to have a stronger command of the language, which helped them detect inconsistencies and errors in speech. 

As shown in Fig. \ref{Fig9}(d), there is no significant correlation between human audiovisual deepfake detection and their IT skill level.
Participants may have overestimated their IT skill level, leading to the result that the human ability to identify authenticity did not improve with increasing IT skill level as expected.

Overall, t-test analysis showed that detection accuracy differed significantly between participants of different age groups and genders, while there were no significant differences in detection accuracy between different IT skill levels or between native and non-native English-speaking participants. Notably, participants showed improved performance in Phase II regardless of grouping, as shown in Fig. \ref{Fig9}. The reasons have been discussed in the previous section.

\subsection{Accuracy vs. Confidence}

The study further analyzes the correlation between accuracy and confidence, as this insight is valuable for assessing participants' 
metacognitive awareness and decision-making. To estimate whether participants correctly perceived their ability to detect audiovisual deepfakes, we compared their actual detection accuracy with self-reported confidence across all videos. Surprisingly, we found that participant’s self-reported confidence in their answers did not align with their accuracy in detecting audiovisual deepfake videos. Despite expressing high confidence, their actual accuracy was low. As shown in Fig. \ref{Fig11}, most participants reported average confidence that was higher than their average accuracy.  
The average confidence level across all participants was 77.60\%, while the average accuracy was 65.64\%. This difference means that humans’ overconfidence in deepfake detection will cause deepfakes to bring greater harm to human society.

People tend to overestimate their ability to identify fake videos, which can lead to false assurances about the reliability of their judgments, poor decision-making, and increased susceptibility to misinformation. Several psychological and cognitive factors may contribute to this overestimation. People often fall victim to cognitive biases, such as the Dunning-Kruger effect \cite{R58}, which often causes people to overestimate their skills in tasks for which they are not strong. Another factor may be that participants interpret their familiarity with certain faces or voices as evidence of authenticity \cite{R59}; this reliance may increase their confidence even if deepfakes manipulate cues they believe to be reliable. Likewise, some people believe that their exposure to the media or prior experience gives them control over discerning real from fake. This illusion of control \cite{R60} may lead to an overestimation of one's own detection capabilities.

\subsection{Detection Intervention}

\begin{figure}[!b]
\vspace{-4mm}
\centering
\includegraphics[width=3.0in]{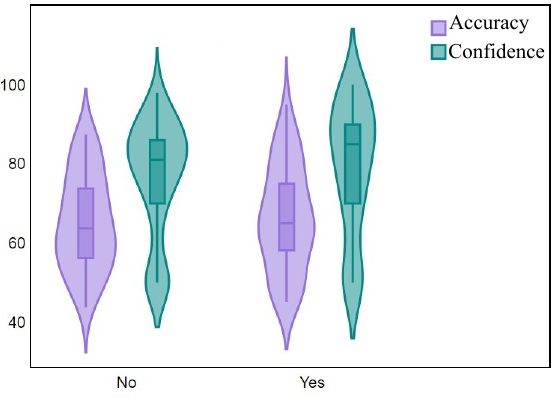}
\caption{Distribution of confidence and accuracy in audiovisual deepfake detection for two participant groups (forewarned vs. not forewarned). The shape and width of the violin body provide insight into the density of data points within each group, while in the middle of each density curve is a small boxplot showing the median (dash) and the first and third quartiles at either end.}
\label{Fig12}
\end{figure}

At first, audiovisual deepfakes may appear convincing; however, AI-generated content can exhibit artifacts, such as irregular shadows, hair anomalies, speech artifacts, glitches, or inconsistencies. An effective approach to enhance participants' performance in detecting audiovisual deepfakes is raising awareness about these deceptive techniques and their adverse effects. Surprisingly, our investigation shows that forewarning of these artifacts and the potential adverse effects of audiovisual deepfake technology did not have any significant impact on participants' performance. As shown in Fig. \ref{Fig12}, participants who received a warning showed similar performance to those who did not receive a warning. Not only did the two groups of participants have similar median accuracy/confidence, but the distributions were also similar. The similar performance in both scenarios may be due to human visual and auditory perception of content. Humans may have cognitive biases and be overconfident in their ability to detect audiovisual deepfakes. Therefore, even after receiving forewarning, there is little incentive to improve their detection capabilities. Moreover, the participants who are forewarned seems to be more confident in their detection abilities. Their increased confidence in audiovisual deepfake detection may stem from their increased awareness, leading them to actively scrutinize the videos for signs of manipulation.

\subsection{Performance of AI Models}
We evaluated 5 SOTA AI models on the same 40 videos used in the subjective evaluation. All models were trained on the FakeAVCeleb dataset \cite{R12}. The results are shown in Table \ref{table2}.
The LipForensics model achieved an accuracy of 92.50\%, correctly identifying 37 out of 40 samples. In comparison, the AV-Lip-Sync model that considers audiovisual consistency achieved an accuracy of 87.50\% by correctly identifying 35 out of 40 samples. The remaining three models, AV-Lip-Sync++, CNN-Ensemble, and AVTENet, achieved an accuracy of 97.50\%, correctly identifying 39 out of 40 samples. 


\begin{table}[!h]
\caption{Detection accuracy of AI models.}
\centering
\scalebox{0.8}{
\begin{tabular}{|c|c|c|c|c|c|c|}
\hline
\textbf {Model} & \textbf {Class} & \textbf {Precision} & \textbf {Recall} & \textbf {F1-Score} & \textbf {Accuracy (\%)} \\
\hline \hline
\multirow{2}{*}{LipForensics \cite{R30}} & Real & 0.95 & 0.90 & 0.92 & \multirow{2}{*}{92.5}  \\
\cline{2-5}
& Fake & 0.90 & 0.95 & 0.93 & \\ 
\hline
\multirow{2}{*}{AV-Lip-Sync \cite{R15} } & Real & 0.94 & 0.80 & 0.87 & \multirow{2}{*}{87.5} \\
\cline{2-5}
& Fake & 0.83 & 0.95 & 0.88 & \\ 
\hline
\multirow{2}{*}{AV-Lip-Sync+ \cite{R11}} & Real & 1.00 & 0.95 & 0.97 & \multirow{2}{*}{97.5}  \\
\cline{2-5}
& Fake & 0.95 & 1.00 & 0.98  & \\ 
\hline
\multirow{2}{*}{CNN-Ensemble \cite{R13}} & Real & 0.95 & 1.00 & 0.98 & \multirow{2}{*}{97.5}   \\
\cline{2-5}
& Fake & 1.00 & 0.95 & 0.97 & \\ 
\hline
\multirow{2}{*}{AVTENet  \cite{R14}} & Real & 0.95 & 1.00 & 0.98 & \multirow{2}{*}{97.5} \\
\cline{2-5}
& Fake & 1.00 & 0.95 & 0.97 & \\ 
\hline
\end{tabular}}
\label{table2}
\end{table}

The accuracy of LipForensics and AV-Lip-Sync is relatively low compared to the other three AI models. This may be because the former is a unimodal method and relies only on lip movements, which makes it insufficient for identifying sophisticated audiovisual deepfakes since manipulation may occur in other facial areas besides lips. In contrast, the latter focuses on audiovisual speech-lip synchronization, but its performance may be affected by variations in audio quality or other factors that impact the synchronization. AV-Lip-Sync+, CNN-Ensemble, and AVTENet achieved the highest accuracy because these models adopt a multimodal approach or utilize advanced sophisticated techniques such as audiovisual synchronization, ensemble network, or transformer architecture, which may enhance their ability to identify various deepfakes.

\begin{figure}[!t]
\centering
\includegraphics[width=3.5in]{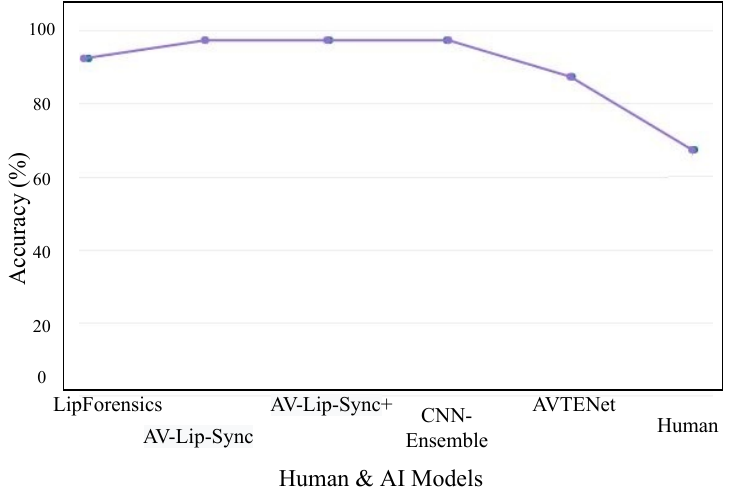}
\caption{Comparison of the performance of humans and 5 AI models in audiovisual deepfake detection.}
\label{Fig13}
\vspace{-6mm}
\end{figure}

\subsection{Human vs. AI Models}

Fig. \ref{Fig13} compares the performance of humans and 5 AI models in detecting audiovisual deepfakes. Human performance was well below expectations and marginally above the chance level. However, all AI models performed significantly better than human participants. Arguably, AI models have demonstrated an impressive ability to detect subtle manipulations in audiovisual content that might escape human perception. We conducted two-sample t-tests to compare human performance against each of the five AI deepfake detection models. As shown in Table \ref{table3}, all AI models demonstrated significantly superior performance in audiovisual deepfake detection compared to humans, with p-values less than 0.05 and large negative t-values indicating a considerable performance gap. The results confirmed that AI models significantly outperformed humans. AI models excel humans at integrating information from multiple modalities, and combining audio and visual modalities in deepfake detection contributes significantly to the effectiveness of AI models. 

\begin{table}[!h]
\caption{Detection Accuracy of Human and AI models.}
\centering
\scalebox{0.9}{
\begin{tabular}{|c|c|c|c|c|}
\hline
\textbf{Method} & \textbf{\begin{tabular}{@{}c@{}}Accuracy \\ (\%)\end{tabular}}   
& \textbf {t-value} & \textbf{\begin{tabular}{@{}c@{}}p-value  \\  
\end{tabular}}  &\textbf{\begin{tabular}{@{}c@{}}Significance  \\ $p<0.05$\end{tabular}} \\
\hline \hline 
Human & 65.64 & - & -  & -  \\
\hline
LipForensics \cite{R30} & 92.50 & 
-23.58  & 4.42 $\times 10^{-45}$ & True \\
\hline
AV-Lip-Sync \cite{R15} & 87.50 & 
-19.54 &  6.40 $\times 10^{-38}$ 
  & True \\
\hline
AV-Lip-Sync+ \cite{R11} & 97.50 & 
-27.33 & 2.18 $\times 10^{-51}$ & True \\
\hline
CNN-Ensemble \cite{R13} & 97.50 & 
-27.38 & 1.91 $\times 10^{-51}$ & True \\
\hline
AVTENet \cite{R14} & 97.50 & 
-27.43 & 2.87 $\times 10^{-51}$ & True \\
\hline
\end{tabular}}
\label{table3}
\vspace{-2mm}
\end{table}

Comparisons between human observers and AI models highlight an important aspect of audiovisual deepfake detection. Despite the adaptability of human perception, AI models in this field consistently outperform humans. Humans were 65.64\% accurate at detecting audiovisual deepfakes, indicating that humans have difficulty identifying forgeries in videos. In comparison, AI models' accuracy was significantly higher, ranging from 87.5\% to 97.5\%. Possible key factors for AI models to perform well are as follows:


\begin{itemize}
    \item Consistency in Decision-Making: AI models follow predefined rules and decision thresholds to ensure absolute consistency in their detection strategies. However, humans suffer from fatigue, cognitive biases, and subjective inconsistencies that affect their decision-making.
    
    \item Feature Extraction Capabilities: Humans rely on visual cues such as facial expressions, lip movements, and tone of voice in deepfake detection, but these features can be convincingly fake. However, AI models use advanced techniques such as CNNs, transformers, ensemble learning, or multimodal features of audio and video to detect subtle manipulations and flag deepfakes that look realistic to human observers.   
    
    \item Scalability and Learning from Data: AI models trained on FakeAVCeleb with variants of manipulated samples can effectively generalize or adapt to different fake generation techniques. In contrast, humans are hindered by limited exposure to deepfakes and cognitive biases. Especially with the rapid development of deepfake generation methods, without the help of automated tools, it is difficult for people's judgment ability to keep up with the times.
    
    \item Fatigue and Bias: Human detection capabilities tend to degrade over time due to fatigue and bias, especially when tasks are repeated. AI models ensure steady performance and no fatigue during large-scale content processing.
\end{itemize}

\section{Limitations}
We are the first to offer insights into the study of human perception of audiovisual deepfakes. In addition to its contribution, this study has several limitations that can be mitigated in future studies to thoroughly explore this field. First, our choice to use a sample size of 40 may not be sufficient to draw general conclusions about human perception in audiovisual deepfake detection tasks. Second, other interventions, such as control situations or providing incentives, can be used to further explore human behavior. Third, to expand the language range, experiments can be conducted in other languages. Fourth, participants can be given multiple samples with different manipulation types to evaluate their performance. 
   
\section{Conclusions and Future Work}
In this paper, we present the results of a subjective study aimed at analyzing human behavior in detecting audiovisual deepfake videos and comparing human performance with that of five AI models. We tested 110 human subjects and five AI models using the same set of 40 manually selected videos (20 real and 20 fake). The findings showed that subjects performed better than chance but generally found detecting deepfakes challenging. They performed marginally better than random chance, even when explicitly informed of the presence of deepfakes. The concern is that deepfake technology has reached a level of reality that could confuse much of the public, especially social media users, while people are overconfident in their ability to determine the authenticity of a video. However, over time, participants seemed to refine their perception and develop a more accurate understanding of their detection capabilities. Additionally, we found that familiarizing participants with the threats posed by deepfakes had no significant impact on their ability to detect deepfakes. Participants' age, gender, and whether they were native speakers of English had an impact on their detection performance, but not their IT skill level. On the other hand, AI models outperformed human participants in detecting forgeries in videos. Furthermore, deepfake detection methods that integrate multiple modalities are more effective than unimodal methods. In order to reduce the harm of deepfakes to human society, it is crucial to develop deepfake detection technology.

This study has important implications for forensic analysis and adaptive countermeasures that improve the accuracy of digital forensic investigations, the attribution and tracking of manipulated content, the validation and verification of audiovisual evidence, and the development of forensic tools. Our findings highlight the need for improved training programs that incorporate AI support into forensic contexts; these insights can help law enforcement and media organizations develop more advanced detection tools and ensure better verification of audiovisual authenticity. Our findings also highlight the importance of adaptive countermeasures, such as real-time monitoring systems, that can dynamically keep pace with the evolution of deepfake technology to maintain the integrity of digital media in critical applications. This research offers a wide range of applications. In forensic analysis, these insights can drive the creation of better tools that combine human intuition with AI models. These tools will strengthen the fight against misinformation and will be valuable for media organizations, law enforcement, and legal professionals to verify audiovisual content.

To further expand and refine our understanding of human perception of audiovisual deepfakes, we can implement a test design that measures subjective performance in a more discreet and unbiased way. For example, the applicability of our findings can be generalized by including a broader and more diverse population. Detailed analysis of the auditory and visual components of audiovisual signals will reveal how individuals perceive, interpret and process audio and visual cues in multimodal deepfake videos. One potential area for future research is analyzing and understanding the cognitive biases that hinder the human ability to identify deepfakes. Furthermore, exploring the impact of the popularity of people in videos (e.g., celebrities or politicians) may provide valuable insights into enhancing human detection capabilities and understanding how prior knowledge affects the perception of corresponding deepfake content.

\bibliographystyle{elsarticle-num}
\bibliography{Reference.bib}

\end{document}